\documentclass[10pt,twocolumn,a4paper]{article}
\usepackage[T1]{fontenc}
\usepackage{mathptmx}
\usepackage[margin=20mm,columnsep=6mm]{geometry}
\usepackage{microtype}
\usepackage{booktabs,longtable,array}
\usepackage{amsmath}
\usepackage{graphicx}
\usepackage{caption}
\usepackage{xcolor}
\usepackage{xurl}
\usepackage{hyperref}
\definecolor{linkink}{HTML}{245B75}
\hypersetup{colorlinks=true,linkcolor=linkink,citecolor=linkink,urlcolor=linkink,
 pdftitle={Can You Undo It? Reversibility Classes for AI Agent Actions},
 pdfauthor={Deepak Waghmare and Shaad Sufi}}
\makeatletter
\renewcommand\section{\@startsection{section}{1}{\z@}{-2.3ex plus -.7ex minus -.2ex}{1.2ex plus .2ex}{\normalfont\large\bfseries\raggedright}}
\renewcommand\subsection{\@startsection{subsection}{2}{\z@}{-2ex plus -.5ex minus -.2ex}{.8ex plus .2ex}{\normalfont\normalsize\bfseries\raggedright}}
\makeatother

\begin{document}
\twocolumn[{
\begin{center}
{\LARGE\bfseries Can You Undo It?\par}
\vspace{3pt}
{\Large\bfseries Reversibility Classes for AI Agent Actions\par}
\vspace{9pt}
{\large Deepak Waghmare\hspace{2.5em}Shaad Sufi\par}
\vspace{3pt}
{\normalsize Two Much Labs, Singapore\par}
\vspace{3pt}
{\small \href{mailto:deepak@twomuchlabs.com}{deepak@twomuchlabs.com}\hspace{2em}\href{mailto:shaad@twomuchlabs.com}{shaad@twomuchlabs.com}\par}
\vspace{4pt}
{\small Research preprint\enspace\textbar\enspace 17 November 2026\par}
\end{center}
\vspace{2pt}
\begin{minipage}{\textwidth}
\small
\textbf{Abstract.}
Tool-using AI agents change files, databases, access rights and public information, yet permission to call a tool does not establish whether a mistake can be undone. We present Reversibility Classes, a declared five-level vocabulary with argument-sensitive rules, conservative defaults and an explicit prior-state capture contract. The pinned implementation and evidence-linked registry cover 536 tools across 20 MCP servers. All 738 rule-case checks conform to declared expectations, while native annotation mapping differs from draft registry labels for 118 of 481 comparable tools. One AI-assisted reviewer approves 148 ratings, 75 matching the registry; this is not independent agreement. On 2,158 recorded runs, the implemented session guard gates the associated action in all 29 attributable attack-positive runs, compared with 18 for a read/write proxy, while adding nine approval points across 1,209 benign runs. Controlled file, Git, SQLite and three-browser experiments show that recovery depends on captured state, later changes, editor storage and recovery route, and expose a Git merge-state exception. The evidence supports making recovery assumptions explicit and testable. It does not establish registry-wide accuracy, reduced production harm or general recovery reliability.
\end{minipage}
\vspace{10pt}
\begin{minipage}{\textwidth}
\centering
\includegraphics[width=.99\textwidth]{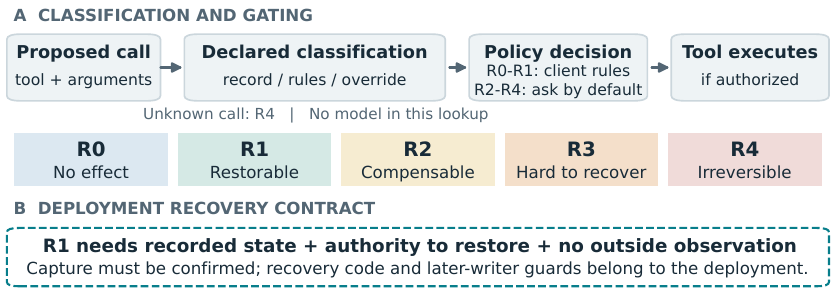}
\captionof{figure}{\textbf{Reversibility is a declared recovery contract.} The classifier maps a proposed call to R0--R4; the gate applies the configured policy. The default hook asks for R2--R4 and leaves client rules in force for R0--R1. A capture-based R1 claim depends on deployment guarantees. Backups, recovery commands and later-writer guards are supplied by the deployment, not executed by the classifier.}
\label{fig:system}
\end{minipage}
\vspace{12pt}
}]

\section{Introduction}

Figure~\ref{fig:system} summarizes the declared mechanism and its deployment boundary. A deleted public message may remain known to its readers. A private file edit may be recoverable from a saved copy, yet replacing the file can erase a colleague's later work. A closed browser tab may recover a textarea in one tested configuration while losing a rich editor in the same browser. These distinctions matter when deciding whether an agent may act autonomously.

Reversibility complements capability restrictions, severity assessment and model confidence. It asks a separate question: given the effects of a successful call and the recovery mechanisms available to the acting principal, what can be restored or compensated? A compensable action can still cause serious harm, consume time or affect people before compensation. Accordingly, we do not use reversibility as a replacement for severity, authorization or domain-specific limits.

This paper contributes (1) a shared five-class vocabulary and machine-readable recovery declarations; (2) a reference classifier, conservative gate and registry of real MCP tools; (3) a measured comparison with MCP annotations and transparent reporting of an assisted review; (4) exploratory replay of recorded agent actions using the implemented session guard; and (5) controlled recovery studies that test specific registry rationales and expose limits of simple undo procedures. The empirical contribution is bounded evidence about these implementations and conditions, rather than a claim to have solved agent safety.

The specification, registry and classifier are available from \href{https://github.com/TwoMuchLabs/reversibility-classes}{TwoMuchLabs/reversibility-classes} \cite{ref1}. This evaluation pins commit \nolinkurl{72faa608}, with package version 0.7.0 and specification version 0.2. Counts refer to that snapshot, not to a continuously changing registry.

\section{Background and related work}

\textbf{Recovery and compensation.} Learned reset policies can prevent reinforcement-learning agents from entering states they cannot reliably leave \cite{ref2}. Relative-reachability penalties address disruption to the states an agent can still reach \cite{ref3}. SagaLLM applies transactional compensation and persistent context to multi-agent planning \cite{ref4}. These precedents establish that recovery, compensation and side-effect control are existing ideas. Our contribution is their operational expression in a shared declaration for heterogeneous tool calls, with explicit treatment of disclosure and deployment assumptions.

\textbf{Runtime gates.} POLAR derives candidate inverse sequences using an ontology and rejects calls below a reversibility threshold \cite{ref5}. It is close in motivation, but evaluates a runtime inverse-search mechanism. Reversibility Classes provide a declared vocabulary, registry and deterministic classification path. The mechanisms could be combined; we do not test that combination or claim superiority to POLAR. AgentTrust combines interception rules with model-based judgment for ambiguous requests \cite{ref6}. The classifier studied here evaluates records and arguments without a model in that classification path, while the creation and review of those records remain judgment-dependent.

\textbf{External effects and evaluation.} ACRFence records irreversible effects to constrain retries after agent checkpoint restoration \cite{ref7}. Its distinction between an internal checkpoint and external consequences is consistent with our refusal to treat a restored local artifact as proof that disclosure has been undone. GAP studies divergence between textual refusal and executed tool actions \cite{ref8}. ToolEmu uses an emulated sandbox to investigate risky actions \cite{ref9}; AgentDojo evaluates prompt injection in stateful tool environments \cite{ref10}; and tau-squared-bench evaluates conversational agents in a dual-control setting \cite{ref11}. Our replay uses recorded runs from the latter two projects. It does not rerun their agents under an intervention.

\textbf{MCP annotations.} The versioned MCP specification provides tool hints including read-only, destructive and open-world properties, and cautions that annotations are untrusted unless their source is trusted \cite{ref12}. These flags are useful inputs but do not directly declare an inverse, restoration authority, compensation cost or retention window. Section 6.2 measures the difference produced by the reference mapping, without treating registry judgments as ground truth.

\section{Classes and the recovery boundary}

A class describes a call succeeding as requested, under a declared environment and principal. Tool malfunction is outside the classification model. The principal is the actor whose existing authority and available recovery interfaces are being assessed. Classes apply to the affected state and consequences, not merely to the visible artifact left after execution.

\textbf{Exact restoration has three requirements.} First, the prior state must be recorded in full, through an accessible version history or a capture made before execution. Second, the principal must already be able to write that state back without another person's intervention. Third, no outside party may have seen, received or acted upon the changed state. An audit log may satisfy the first condition without satisfying either of the others.

\textbf{Disclosure and incidental effects.} The pinned specification classifies intended delivery of content to outsiders as R4 even if a stored message can be deleted. Its present convention classifies state changes with incidental notifications by the recoverability of that state change. This convention creates a scope question: a reviewer who includes every possible notification or observation may assign a higher class. We retain that convention as part of the evaluated specification and expose its consequences rather than treating all disagreement as reviewer error. Where disclosure is actually in scope, a backup cannot undo it.

\textbf{Restoration and compensation.} R1 requires restoration; R2 permits a bounded difference accepted by the principal. The SQLite AUTOINCREMENT experiment makes this distinction concrete: dropping the newly created user table leaves an empty internal sequence table. Compensation cost and acceptable residue require explicit conditions. A nominal refund or delete operation alone does not demonstrate that all consequences are bounded.

The highest applicable effect determines the class, while ordered argument rules select among the explicitly declared cases. Unrecognized calls use R4. Environment-specific lowering requires a documented override; a disclosure cannot be lowered while it remains a disclosure. These mechanisms make assumptions inspectable. They cannot ensure that the assumptions are true in a deployment.

\section{Records, classification and enforcement}

\subsection{Records and argument-sensitive rules}

Each server record links its tool list to source evidence and declares a class and rationale for each tool. Optional fields identify disclosure, conditional rules, compensation, undo windows and review status. The evaluated registry entries retain draft status. A class declared for a tool's broadest effect is the conservative base. Ordered rules can identify narrower argument forms; the first match applies. Missing or unmatched inputs keep the base class unless an explicit rule covers them.

Lowering rules should identify a positive, narrowly defined form rather than the absence of an obvious dangerous keyword. A database call beginning with SELECT is not sufficient evidence of read-only execution: functions and server behavior matter. The initial effect checks test simple accepted queries and filesystem previews; they do not prove that every possible SQL string or argument encoding is safe.

\subsection{Defaults and policy}

The trusted-annotation mapping yields R0 for read-only hints; otherwise it yields R4 for open-world operations, R1 for closed-world non-destructive operations, and R3 for closed-world destructive operations. Missing hint fields use the protocol defaults. No branch yields R2. Explicit registry records take precedence, and an unclassified tool without an applicable trusted mapping remains R4.

The specification's suggested autonomy levels appear in Table~\ref{tab:main1}. In the evaluated hook, \nolinkurl{GatePolicy.decide} returns "ask" for R2-R4 and no decision for R0-R1, leaving existing client permissions in force. Optional configuration can explicitly allow low classes or deny high ones. Thus, "the classifier returned R0" does not itself mean that the deployed client executed a call without approval. Our replay counts the policy's projected approval points with no R2 budget.

\subsection{Deployment overrides and prior-state capture}

A deployment can record a reason for raising a class or lowering a non-disclosure class. A capture-based R1 override declares the state to save, how to restore it and its retention. Capture must be confirmed before the call; failure reverts to the underlying class. Each captured decision must carry its capture reference in the audit record. Capturing a private file cannot justify an override covering payments, other files, access changes, triggered automation or disclosure.

The classifier verifies the declaration and confirmation contract. It does not create the backup, discover all effects, perform a merge or prove that no outsider observed the action. Those responsibilities belong to the deployment. In our experiments, the harness supplies backup storage, rollback commands, conditional writes and conflict detection. We report those mechanisms separately from the unmodified classifier.

\subsection{Session-aware enforcement}

The pinned implementation includes session tracking and \nolinkurl{session_guard}, so sequence-aware enforcement is not solely proposed future work. The guard raises an egress-capable call to R4 after the session has read data unless every destination is allow-listed. The replay invokes this implementation for its reviewed session arm. The web-read adapter uses the registry's R0 fetch record with egress checks enabled, an empty destination allow-list and fresh session state per trace. Other benchmark functions retain the analyst's mappings. We test this configuration, not every sequence rule, integration or tamper-resistance property of the hook.

\section{Registry and study design}

The pinned registry contains 536 tools across 20 servers, including reference and vendor MCP implementations. Its base-label distribution is R0: 333 (62.1\%), R1: 31 (5.8\%), R2: 18 (3.4\%), R3: 39 (7.3\%) and R4: 115 (21.5\%). Of the 203 non-R0 entries, 115 are labeled R4 (56.7\%). These are descriptive counts of draft judgments, not measured rates of irreversible behavior in deployed tools.

The registry was drafted and audited with AI assistance. Earlier audit passes motivated corrections to SQL, disclosure and undo assumptions. We do not treat those passes as independent validation or combine their historical correction counts with the present experiments. The completed study uses the following distinct forms of evidence; their denominators must not be added into a single "success rate."

All new effect experiments use synthetic local fixtures. Tests were selected to probe specific recovery claims rather than sampled to estimate registry-wide accuracy. Operations involving later writers are serialized; they do not test races during rollback. Recorded traces were collected by the benchmark projects, not generated by this study. The analysis is exploratory, without preregistration or independence-based significance estimates.

\section{Results}

\subsection{Assisted human review}

Using seed 20261115, the study selected 150 tools stratified by registry class: 34 R0, 31 R1, 18 R2, 33 R3 and 34 R4. All available R1 and R2 entries were included. The reviewer, coauthor Shaad Sufi, completed a decision for every item: 148 class ratings and two explicit holds for insufficient implementation evidence (Stripe recipient setup and currency conversion).

The first nine ratings were supplied interactively by the reviewer with explanations; one of these followed an AI recommendation. For the remaining 139 rated items, the reviewer approved AI-proposed ratings in batches. The two holds were also approved. The resulting rating distribution was R0: 29, R1: 2, R2: 26, R3: 26 and R4: 65. Of the 148 ratings, 75 matched the draft registry exactly (50.7\%), 67 were higher and six lower.

This is a completed assisted review, not a blinded inter-annotator study. The review record is preserved, including suggestions and rationales. No independent second or third rating set was submitted, so we do not report Cohen's kappa, Krippendorff's alpha or a claim of independent agreement. The original proposal for a three-person agreement experiment has not been fulfilled by this procedure.

Differences include assumptions about outside observation, who can perform recovery and whether a possible failure must be included. Some suggestions also contained analyst errors: an R0 recommendation for \nolinkurl{browser_find} missed its output-filename option. Consequently, the 50.7\% match rate is neither an accuracy estimate nor proof that the taxonomy is invalid. It identifies specification and review-process questions. We retain the approved judgments as collected instead of silently editing them after later experiments.

\subsection{Native annotations and registry labels}

\begin{figure*}[t]
\centering\includegraphics[width=\textwidth]{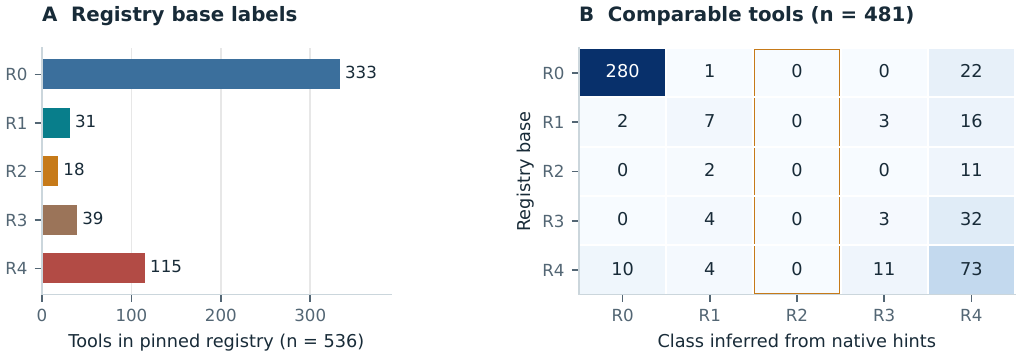}
\caption{\textbf{Native hints lose distinctions expressed by the registry.} (A) Base-label composition of 536 draft registry entries. (B) Annotation mapping for 481 comparable tools, with raw counts in each cell. The outlined R2 column is empty by construction. The 55 unmatched tools are excluded from the matrix. Disagreement with draft labels is not an accuracy estimate; full counts are in Table~\ref{tab:main3}.}\label{fig:hints}
\end{figure*}

Verified comparable hint evidence was extracted for 481 of 536 tools (89.7\%) across 14 servers. The 55 unmatched tools comprise 29 archived-server tools, 15 hosted Stripe tools, seven Terraform tools and four GitHub tools. Unmatched means that the study did not verify a comparable object from its pinned sources; it does not mean a deployed server lacks annotations.

Figure~\ref{fig:hints} shows the registry composition and annotation cross-tabulation. There were 363 exact matches and 118 differences (24.5\% of comparable tools): the hint mapping was higher in 85 and lower in 33. All 13 comparable R2 entries differed because the mapping cannot express R2. At the study's R2 approval threshold, 61 tools crossed the boundary: hints placed 20 below it where registry labels required approval, and 41 above it where registry labels did not. These are policy differences against draft judgments, not verified false-positive or false-negative rates.

\subsection{Recorded policy replay}

\begin{figure*}[t]
\centering\includegraphics[width=\textwidth]{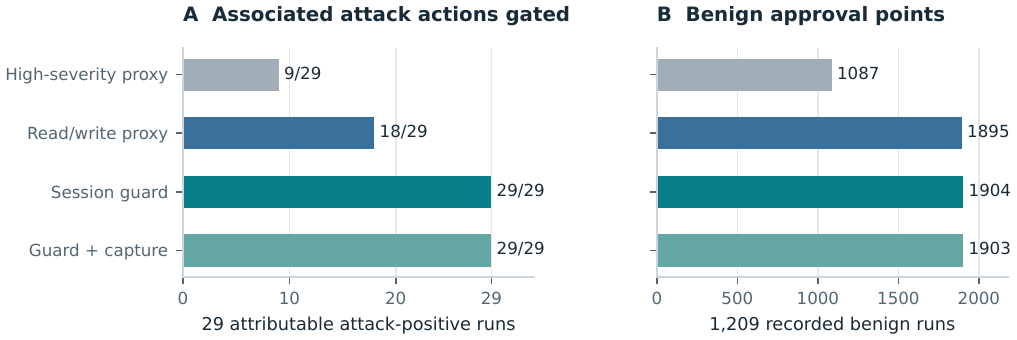}
\caption{\textbf{A measured replay tradeoff, with separate denominators.} The implemented session guard covers 29 associated actions versus 18 for the read/write proxy, at 1,904 versus 1,895 benign approval points. All 11 additional covered cases depend on web-exposure modeling. Bars start at zero. The selected 29-run attribution subset and 1,209 benign runs are not intervention outcomes; all seven policy arms appear in Table~\ref{tab:main4}.}\label{fig:replay}
\end{figure*}

The dataset contains 1,046 AgentDojo runs and 1,112 tau-squared-bench runs: 2,158 in total, covering seven domains and one agent-model cohort, Claude 3.7 Sonnet. There are 1,209 benign runs with 8,761 agent calls and 949 injected runs with 3,315 calls. The study maps 106 distinct benchmark tool names. These benchmark functions are not the 536 registry-native MCP tools, so the adapter is part of the method rather than evidence that every registry label was exercised.

We compare a read/write proxy, a high-severity proxy and the implemented session-aware gate. Earlier mappings that classed every web visit R4, a sensitivity arm treating web visits as ordinary reads, and capture-enabled variants are retained for transparency. An approval point is a recorded call where the projected policy asks; the replay continues counting later calls even though a real refusal might have changed the rest of the trace.

Figure~\ref{fig:replay} separates attack-action coverage from benign approval burden.

\textbf{Attack-action coverage.} Of 47 runs with a positive benchmark attack-goal label, 29 permit association with an action through an explicit benchmark destination marker or a sole candidate action under the reviewed predicate. The other 18 positives are excluded from action attribution. The implemented session guard gates the associated action in 29/29, versus 18/29 for the read/write proxy and 9/29 for the high-severity proxy. All 11 additional cases relative to the read/write proxy depend on the treatment of web retrieval as a possible exposure channel. Ordinary-web-read sensitivity returns coverage to 18/29. This is a specific modeling result, not evidence of a general 100\% attack-prevention rate.

\textbf{Approval burden.} The implemented session guard adds nine benign approval points over the read/write proxy (1,904 versus 1,895, an increase of 0.47\%). Per benign run, the means are 1.5749 and 1.5674. Paired trace counts are identical in 1,196 runs, higher under the guard in nine and lower in four. Thus, the hypothesis that this arm lowers benign approval burden is not supported. The number of original benign successes with no projected prompt is 119 under the guard and 118 under the read/write proxy; neither number measures success after a user accepts or refuses a prompt.

\textbf{Capture arm.} Of 59 recorded file appends, 15 meet the simulated private-capture conditions: one benign and 14 injected. All 15 saved fixtures restore exactly under the simulated backing store and authority assumptions. Capture removes one benign and 14 injected approval points from the corresponding non-capture arm. These are simulated deployment interventions on eligible file appends, not historical backups or executed database recovery. No model behavior, approval decision or prevented harm is inferred after a hypothetical pause.

\subsection{Rule checks and initial effects}

The rule suite contains 738 rule-case combinations covering 633 distinct tool/argument inputs, 58 lowering rules and 40 tools. Expected classes and selected-rule identities match in all 738 checks. The generated cases include ordinary accepted values, missing and wrong-type arguments, nested fields and fixed URL/SQL boundary examples. Because the expectations are derived from the same declared rules, this result demonstrates implementation conformance on the tested inputs, not independent validation of the rules. The 62 upstream conformance tests also pass.

Nine initial effect observations exercise two rule families. Two filesystem previews preserve the original bytes. Four actual filesystem edits change the file and then restore its prior bytes using the harness capture. Three queries on a disposable PostgreSQL 14.20 instance return the expected values 1, 2 and 11 and preserve the fixture rows. The filesystem experiment executes the pinned upstream edit helper; the PostgreSQL probe tests database effects directly rather than starting the original MCP server. These checks are deliberately narrow and do not establish adversarial robustness or safety of arbitrary SQL.

\subsection{File recovery with later changes}

\begin{figure*}[t]
\centering\includegraphics[width=\textwidth]{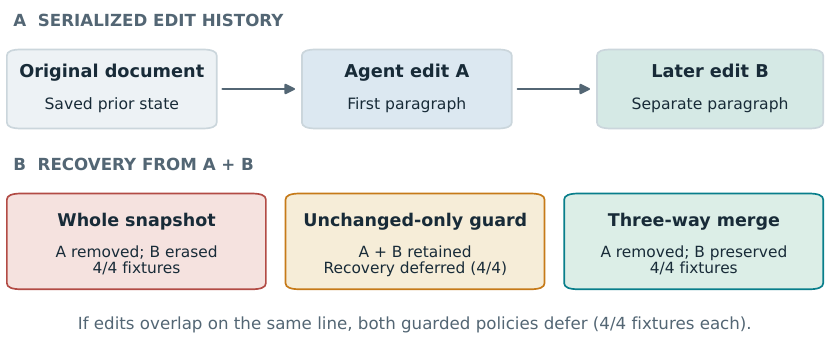}
\caption{\textbf{Restoring the starting state can destroy later work.} The schematic corresponds to the separate-paragraph fixture, repeated across four encoding/line-ending variants. Whole-file rollback erases the later change; a three-way merge preserves it; a conservative guard defers. The classifier supplies the capture contract, while the harness supplies all recovery algorithms. All scenarios, including unavailable capture, appear in Table~\ref{tab:main5}.}\label{fig:later}
\end{figure*}

We cross four fixtures (ASCII or Unicode, each with LF or CRLF line endings), four scenarios and three recovery policies, producing 48 observations. The policies are unconditional whole-file snapshot restoration, restoration only if the file is unchanged since the agent's edit, and a three-way text merge using the captured prior state. The classifier supplies the preimage confirmation and fallback contract; the harness supplies all three recovery algorithms.

Figure~\ref{fig:later} illustrates the separate-paragraph case. Restoring the starting bytes and preserving later legitimate work are different endpoints. A snapshot satisfies the first while failing the second in both later-writer scenarios. The three-way procedure succeeds on the separate-paragraph fixtures but defers overlapping edits. These observations support conditional recovery and conflict detection. They do not show that the classifier itself implements merging, or that the procedure handles arbitrary files, simultaneous writes, metadata or external observers.

\subsection{Git and SQLite recovery}

\begin{figure*}[t]
\centering\includegraphics[width=\textwidth]{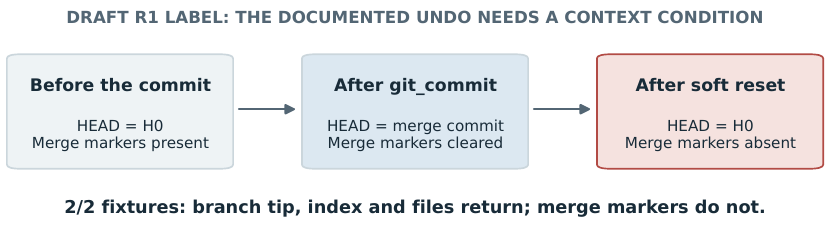}
\caption{\textbf{An exception to an unconditional Git restoration claim.} For both pending-merge fixtures, the documented soft reset restores the measured branch tip, index and files but not the pending-merge markers. This qualifies the draft R1 label for this recovery route; it does not prove that every alternative recovery procedure fails. Full state records and scenario outcomes accompany Tables~\ref{tab:main6} and~\ref{tab:git-scenarios}.}\label{fig:git}
\end{figure*}

We invoke six unmodified upstream tools through local MCP stdio servers. They span all five registry classes. Sixteen scenarios, each with two text fixtures, produce 32 observations. A complete repeat from fresh fixtures reproduces the recorded outcomes and state hashes. This is 16 scenario types, not 64 independent tests of generalization.

For Git, the measured state includes branch tip, staging-area contents, working-file contents and pending-merge markers. It excludes reflogs, dangling objects and index timestamps. For SQLite, it includes logical schema and rows, not database-file layout. This explicit boundary prevents "exact" from being confused with restoration of every physical byte or all historical effects.

Figure~\ref{fig:git} visualizes the pending-merge exception. The AUTOINCREMENT residue and partial-staging behavior reproduce concrete registry rationales. The pending-merge result qualifies the draft R1 label for \nolinkurl{git_commit}: its documented soft reset does not restore all measured merge state in that context. This is evidence against an unconditional recovery claim, not proof that no alternative recovery procedure exists. The later-writer cases again show why return to the initial state is insufficient: dropping a table can erase a new row while making the database look exactly as it did before the agent acted.

\subsection{Cross-browser draft recovery}

\begin{figure*}[t]
\centering\includegraphics[width=\textwidth]{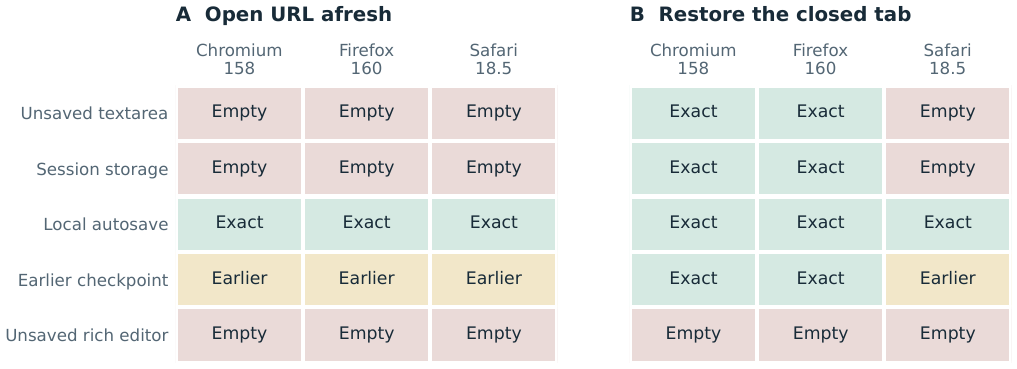}
\caption{\textbf{Recoverability depends on editor storage and recovery route.} Each cell represents two synthetic drafts, each repeated once. Exact means the final text returns; Earlier means only the first edit returns; Empty means no text returns through the tested route. Browser builds, profiles and automation methods differ, so the comparison is not an isolated engine effect. Restart results and unmeasured Safari conditions are explicit in Table~\ref{tab:main8}.}\label{fig:browser}
\end{figure*}

We use the same local form HTML with two synthetic drafts (ASCII and Unicode). Four textarea designs have no persistence, session storage, local autosave on every input, or a local checkpoint containing only the first edit. A fifth fixture is an unsaved contenteditable rich editor. Exact recovery means equality with the final visible text; "earlier" means only the first edit returns; "empty" means neither draft appears in the tested editor.

Chromium 158.0.8127.18 (headless Chrome for Testing through pinned Playwright MCP) and shipping Firefox 160.0 (headless through geckodriver 0.37.1) each have 46 cases: 40 textarea cases over five routes and six rich-editor cases over three routes. Each set is repeated, giving 92 observations per browser. Firefox reproduces every Chromium outcome. The five textarea routes are leave-open control, fresh URL, native closed-tab restore, same-profile browser restart and isolated-profile browser restart. Rich editors are tested only for leave-open, fresh URL and native restore.

Safari 18.5 is tested through native desktop UI with a dedicated study window because WebDriver remote automation is disabled. The local observer initializes each unique fixture once using browser edit operations, then only observes on subsequent loads. Native restoration uses History > Reopen Last Closed Tab and verifies the recovered URL in the UI. Twenty cases cover two recovery routes, five editors and two drafts; a repeat uses distinct URLs and keys in the same desktop profile and loopback origin. All 40 pre-close checks match the intended latest text and storage. Safari has no separate leave-open trial or whole-browser restart measurement.

Figure~\ref{fig:browser} compares the common tab-close routes. The same "close tab" operation has different measured outcomes depending on editor storage and recovery route. Native tab restore recovers unsaved textareas in the tested Chromium and Firefox configurations but not in the tested Safari configuration. Unsaved rich editors return empty on both recovery routes in all three. The first-edit checkpoint is insufficient to restore the final draft wherever the browser does not independently restore that final state.

Browser versions, profile configuration, headless versus desktop execution, edit initialization and automation methods differ. Therefore, the observed difference cannot be attributed solely to browser engine. These experiments do not test crash recovery, server-side drafts, logins or network submissions. Chromium uses an extension to restore the exact closed URL through the sessions API, and Firefox uses the corresponding sessions API. Safari uses native UI. Temporary profiles and synthetic fixtures constrain the inference.

The unchanged registry labels \nolinkurl{browser_tabs} close and \nolinkurl{browser_close} R3; the assisted reviewer rated the browser-close case R4. The results explain why a universal label from a tool name is insufficient. They do not establish that every close is R4 or that an empty field proves all recovery mechanisms impossible. A deployment claiming R1 needs a verified path that restores the relevant draft under its actual conditions.

\section{Discussion}

\textbf{A practical vocabulary with executable rules.} The implementation can consistently apply declared classes, conditional rules and capture fallback on the tested cases. The hint comparison demonstrates a representational difference: the current mapping cannot express compensation, and it changes approval outcomes for 61 comparable entries. Whether each resulting label is substantively correct still requires evidence about effects and recovery.

\textbf{Recovery declarations should be falsifiable.} A record should identify the affected state, the principal's authority, saved information, the recovery action and the conditions under which it is valid. The SQLite residue confirms one published rationale; the Git merge case identifies a missing condition. Both are useful outcomes for an evidence-linked registry. An evaluation that reports only successful recoveries would miss the latter.

\textbf{Later work is a separate invariant.} A restoration check should measure both removal of the mistaken effect and preservation of subsequent legitimate changes. The file and database experiments demonstrate that equality with the starting state can coexist with data loss. Guards and selective recovery can preserve later work, sometimes by deferring to a human. These are deployment design requirements supported by the fixtures, not a universal recovery algorithm supplied by the classifier.

\textbf{Classification and recovery have distinct roles.} The classifier can require confirmed capture and fall back when capture is absent. It cannot prove that the saved state is complete, that an application has no external effects or that a recovery procedure remains safe after another writer acts. The study's merges and conditional SQL updates should not be attributed to the released classification package.

\textbf{Replay establishes a conditional tradeoff.} With the stated adapters and egress configuration, the implemented gate covers 11 more attributable attack-associated actions than the read/write proxy, at nine additional benign approval points. The entire coverage difference is sensitive to web-read modeling. This is evidence for examining exposure through apparently read-only tools, not for broad superiority to other agent defenses. The replay gives no production harm, latency, financial loss, user-acceptance or post-intervention utility estimate.

\textbf{Organizational utility is a hypothesis.} Explicit records may help operators discuss permissions, backups and approval requirements consistently. Logging can provide inspectable evidence of the rule and override used for an action. Neither the current review nor the experiments establish that different organizations interpret the vocabulary identically, that users approve appropriately, or that the records satisfy a particular regulatory duty.

\section{Limitations}

\textbf{Human validity.} The completed review is a single coauthor's AI-assisted assessment. Batch approval, exposure to suggestions and unverified assumptions limit independence. The two explicit holds preserve uncertainty. A future blinded multi-rater study would need a fixed effect boundary, sufficient source evidence and ratings collected before discussion. Such a study is not represented as completed here.

\textbf{Semantic scope.} Passing rule checks cannot validate the policy oracle from which they were generated. The nine initial effect probes cover only two rule families. Neither they nor the controlled later studies establish robust classification of arbitrary code, SQL, scripts or hidden downstream automation. Registry records can become stale when tools or deployments change.

\textbf{Sampling and causality.} Tools and recovery scenarios were deliberately selected. Text variants and repeats are correlated, and browser arms differ in more than engine. There is no estimate of recovery frequency across web applications or tools. The trace cohort contains one model and one injected condition; action attribution is selective and analyst-defined. Mappings, baselines and capture eligibility require judgment. No randomized deployment comparison or intervention on agent behavior was performed.

\textbf{State and concurrency.} Local byte equality, Git's measured state and SQLite logical equality are different endpoints. External observation is excluded by the fixture design rather than measured across real users. Later writers are serialized, leaving races between verification and rollback untested. Simultaneous writes, crashes, metadata, retention expiry, permissions changes, access revocation and production integrations remain outside the demonstrated results.

\textbf{Taxonomy scope.} The boundary between intended disclosure and incidental notification requires clearer operational guidance. R3 combines recovery requiring a human with categorical treatment of access and availability effects. Browser restoration shows that a single base label may hide materially different routes. More explicit context fields or preconditions could improve records, but the present experiments do not validate a revised taxonomy.

\textbf{Future evaluation.} The next tests should prioritize independent annotation with explicit assumptions, state-aware registry corrections for documented exceptions, safe recovery under concurrent changes, and live policy trials measuring task completion, approval burden and harm-related endpoints. Any broader autonomy claim should depend on those measurements rather than extrapolation from deterministic fixtures.

\section{Conclusion}

Reversibility Classes provide a machine-readable way to distinguish exact restoration, compensation, difficult recovery and irreversible effects before a tool call. The pinned classifier follows its declared rules on the tested suite, and controlled experiments confirm several recovery rationales while exposing a Git merge-state exception and browser-dependent restoration behavior. Recorded replay shows a specific coverage improvement from exposure-aware gating with a small increase, not a reduction, in benign approval points.

\begin{quote}\itshape The supported central claim is that agent permission policies can make recovery assumptions explicit and testable, and that those assumptions materially affect whether an action can be undone. The present evidence does not establish that all registry labels are correct, that recovery is universally reliable or that deploying the system reduces real-world harm.\end{quote}

For deployments, "there is an undo command" is an insufficient justification for autonomy. A defensible restoration claim needs the relevant prior state, authority to restore it, a verified recovery path, protection of later work and a boundary that excludes irreversible external effects.

\section{Reproducibility and disclosure}

The accompanying bundle contains the LaTeX manuscript source, compilation instructions, original measurement files, replay mappings, reviewer decisions, protocols and verification scripts. The manuscript metrics are checked directly against the saved CSV files. Each study retains its own reproduction instructions and source pins. The companion data archive contains all five study directories and a SHA-256 manifest of its members. Appendix~\ref{app:data} provides additional data tables and an inventory. The artifact manifest records file hashes; readers can distinguish frozen observations from later edits.

Core sources are the classifier commit listed in Section 1, upstream MCP reference servers at \nolinkurl{51b6382a}, archived servers at \nolinkurl{9be4674d} and Playwright MCP at \nolinkurl{ab6d67cb}. The browser package used Playwright \nolinkurl{1.64.0-alpha-1790635538000} with the explicitly selected Chromium build reported above. Complete environment details and per-study pins are retained in the scripts and protocols. Reproduction requires the appropriate browser/runtime dependencies; public source archives and full benchmark traces are fetched separately rather than embedded in this bundle.

\textbf{AI-use disclosure.} AI assistance was used in registry drafting and auditing, experimental implementation, data analysis, manuscript editing and the human review process described in Section 6.1. Human approval of suggestions is not presented as independent annotation. Synthetic fixture text is used for the effect studies; the exported replay representation retains identifiers, action metadata and labels rather than underlying user-message content. The authors remain responsible for interpretation and final submission.

\clearpage\onecolumn

\clearpage

\appendix
\renewcommand{\thetable}{A\arabic{table}}

\section{Data supplement and evidence inventory}\label{app:data}

This appendix reports additional disaggregated results derived from the saved observations. It does not introduce new trials. The companion archive contains the complete packaged measurement tables, fixture states, reviewer records, protocols and scripts under \path{data/}. Primary observations, repetitions, derived summaries and blank review templates have distinct roles. A repeated deterministic fixture is not an independent population sample. Tables A1--A9 retain the complete main-study results; Tables A10--A14 expose additional breakdowns and the file inventory.
\subsection{Complete main-study result tables}
The full numerical tables are retained here to keep the main narrative focused on the mechanism and measured comparisons. The table labels are stable cross-references from the main text. Draft labels, assisted ratings, projected approvals and executed effects remain distinct forms of evidence.
\par\noindent\begin{minipage}{\textwidth}
\small\setlength{\tabcolsep}{4pt}\renewcommand{\arraystretch}{1.22}
\captionof{table}[]{Specification classes and suggested policy. An R4 action may be explicitly authorized; the label prohibits autonomous execution under this policy. The hook implementation's defaults are described separately in Section 4.}\label{tab:main1}
\begin{tabular}{@{}>{\raggedright\arraybackslash}p{\dimexpr 0.08\textwidth-1.5\tabcolsep\relax}>{\raggedright\arraybackslash}p{\dimexpr 0.17\textwidth-1.5\tabcolsep\relax}>{\raggedright\arraybackslash}p{\dimexpr 0.48\textwidth-1.5\tabcolsep\relax}>{\raggedright\arraybackslash}p{\dimexpr 0.27\textwidth-1.5\tabcolsep\relax}@{}}
\toprule

\textbf{Class} & \textbf{Name} & \textbf{Recovery criterion} & \textbf{Default autonomy policy} \\

\midrule

R0 & No effect & No external state change within the declared effect boundary. & Act autonomously. \\
R1 & Restorable & Prior state is recorded, writable by the principal and unobserved outside its control. & Act, then notify. \\
R2 & Compensable & A known further action reaches an accepted equivalent state at a bounded, knowable cost. & Ask, or use a pre-approved budget. \\
R3 & Hard to recover & Recovery requires a human, outside process or unbounded effort; includes access, credentials, availability or physical-state changes. & Ask for each action. \\
R4 & Irreversible & Disclosure, unrecoverable destruction or another consequence that cannot be recalled or compensated. & Never autonomous. \\

\bottomrule
\end{tabular}
\end{minipage}\par\vspace{11pt}
\par\noindent\begin{minipage}{\textwidth}
\small\setlength{\tabcolsep}{4pt}\renewcommand{\arraystretch}{1.22}
\captionof{table}[]{Evidence inventory. Text variants and repeated runs measure repeatability, not independent samples of a tool population. Detailed inputs, scripts and measurements accompany the manuscript.}\label{tab:main2}
\begin{tabular}{@{}>{\raggedright\arraybackslash}p{\dimexpr 0.19\textwidth-1.3333333333333333\tabcolsep\relax}>{\raggedright\arraybackslash}p{\dimexpr 0.34\textwidth-1.3333333333333333\tabcolsep\relax}>{\raggedright\arraybackslash}p{\dimexpr 0.47\textwidth-1.3333333333333333\tabcolsep\relax}@{}}
\toprule

\textbf{Study} & \textbf{Unit and coverage} & \textbf{Question addressed} \\

\midrule

Assisted review & 150 decisions; 148 ratings; one reviewer & Where do approved judgments differ from draft labels? \\
Hint comparison & 536 entries; 481 comparable hints & What information is lost under the annotation mapping? \\
Recorded replay & 2,158 runs; 12,076 proposed calls & Which recorded actions would cause a projected pause? \\
Rule and initial effects & 738 rule checks; 62 upstream tests; 9 effect observations & Does implementation follow declarations, and do selected effects match? \\
Private-file recovery & 48 policy-scenario-fixture observations & Can rollback preserve later changes? \\
Git and SQLite & 6 tools; 16 scenarios x 2 fixtures, repeated & Do named undo actions restore measured state? \\
Browser recovery & Chromium 46, Firefox 46, Safari 20 cases; each repeated & Which drafts survive each available recovery route? \\

\bottomrule
\end{tabular}
\end{minipage}\par\vspace{11pt}
\par\noindent\begin{minipage}{\textwidth}
\small\setlength{\tabcolsep}{4pt}\renewcommand{\arraystretch}{1.22}
\captionof{table}[]{Base registry class against the reference trusted-annotation mapping. Rows total 536; comparable cells total 481. Conditional arguments and deployment overrides are outside this comparison.}\label{tab:main3}
\begin{tabular}{@{}>{\raggedright\arraybackslash}p{\dimexpr 0.22\textwidth-1.7142857142857142\tabcolsep\relax}>{\raggedright\arraybackslash}p{\dimexpr 0.13\textwidth-1.7142857142857142\tabcolsep\relax}>{\raggedright\arraybackslash}p{\dimexpr 0.13\textwidth-1.7142857142857142\tabcolsep\relax}>{\raggedright\arraybackslash}p{\dimexpr 0.13\textwidth-1.7142857142857142\tabcolsep\relax}>{\raggedright\arraybackslash}p{\dimexpr 0.13\textwidth-1.7142857142857142\tabcolsep\relax}>{\raggedright\arraybackslash}p{\dimexpr 0.13\textwidth-1.7142857142857142\tabcolsep\relax}>{\raggedright\arraybackslash}p{\dimexpr 0.13\textwidth-1.7142857142857142\tabcolsep\relax}@{}}
\toprule

\textbf{Registry base} & \textbf{Hint R0} & \textbf{Hint R1} & \textbf{Hint R2} & \textbf{Hint R3} & \textbf{Hint R4} & \textbf{Missing} \\

\midrule

R0 & 280 & 1 & 0 & 0 & 22 & 30 \\
R1 & 2 & 7 & 0 & 3 & 16 & 3 \\
R2 & 0 & 2 & 0 & 0 & 11 & 5 \\
R3 & 0 & 4 & 0 & 3 & 32 & 0 \\
R4 & 10 & 4 & 0 & 11 & 73 & 17 \\

\bottomrule
\end{tabular}
\end{minipage}\par\vspace{11pt}
\par\noindent\begin{minipage}{\textwidth}
\small\setlength{\tabcolsep}{4pt}\renewcommand{\arraystretch}{1.22}
\captionof{table}[]{Recorded-run projections. Benign denominator: 1,209 runs. Attack-action coverage denominator: the selective 29-run attribution subset, not all 949 injected runs. The final column refers to the 770 originally successful benign runs and does not estimate completion after an intervention.}\label{tab:main4}
\begin{tabular}{@{}>{\raggedright\arraybackslash}p{\dimexpr 0.34\textwidth-1.6\tabcolsep\relax}>{\raggedright\arraybackslash}p{\dimexpr 0.18\textwidth-1.6\tabcolsep\relax}>{\raggedright\arraybackslash}p{\dimexpr 0.17\textwidth-1.6\tabcolsep\relax}>{\raggedright\arraybackslash}p{\dimexpr 0.15\textwidth-1.6\tabcolsep\relax}>{\raggedright\arraybackslash}p{\dimexpr 0.16\textwidth-1.6\tabcolsep\relax}@{}}
\toprule

\textbf{Replay policy} & \textbf{Benign approval points} & \textbf{Injected approval points} & \textbf{Associated actions gated / 29} & \textbf{Prompt-free original benign successes} \\

\midrule

High-severity proxy & 1,087 & 234 & 9 & 312 \\
Read/write proxy & 1,895 & 744 & 18 & 118 \\
Implemented session guard & 1,904 & 765 & 29 & 119 \\
Session guard + capture & 1,903 & 751 & 29 & 120 \\
Earlier all-web-R4 mapping & 1,911 & 800 & 29 & 118 \\
Earlier mapping + capture & 1,910 & 786 & 29 & 119 \\
Ordinary-web-read sensitivity & 1,891 & 688 & 18 & 119 \\

\bottomrule
\end{tabular}
\end{minipage}\par\vspace{11pt}
\par\noindent\begin{minipage}{\textwidth}
\small\setlength{\tabcolsep}{4pt}\renewcommand{\arraystretch}{1.22}
\captionof{table}[]{File recovery outcomes. The 12 capture-failure observations repeat four unique fixtures across three policies. Deferral preserves the current file; it does not complete recovery.}\label{tab:main5}
\begin{tabular}{@{}>{\raggedright\arraybackslash}p{\dimexpr 0.25\textwidth-1.5\tabcolsep\relax}>{\raggedright\arraybackslash}p{\dimexpr 0.25\textwidth-1.5\tabcolsep\relax}>{\raggedright\arraybackslash}p{\dimexpr 0.25\textwidth-1.5\tabcolsep\relax}>{\raggedright\arraybackslash}p{\dimexpr 0.25\textwidth-1.5\tabcolsep\relax}@{}}
\toprule

\textbf{Scenario} & \textbf{Whole snapshot} & \textbf{Unchanged-only guard} & \textbf{Three-way merge} \\

\midrule

No later edit & Restores original bytes, 4/4 & Restores original bytes, 4/4 & Restores original bytes, 4/4 \\
Later edit to separate paragraph & Erases later work, 4/4 & Defers; preserves later work, 4/4 & Removes agent edit and preserves later work, 4/4 \\
Later edit to same line & Erases later work, 4/4 & Defers; preserves later work, 4/4 & Defers; preserves later work, 4/4 \\
Capture unavailable & Fallback R4; edit withheld, 4/4 & Fallback R4; edit withheld, 4/4 & Fallback R4; edit withheld, 4/4 \\

\bottomrule
\end{tabular}
\end{minipage}\par\vspace{11pt}
\par\noindent\begin{minipage}{\textwidth}
\small\setlength{\tabcolsep}{4pt}\renewcommand{\arraystretch}{1.22}
\captionof{table}[]{Six tools, 16 scenarios and 32 primary observations. Recovery commands, captured values and guards are study-harness mechanisms, not additional capabilities of the classifier.}\label{tab:main6}
\begin{tabular}{@{}>{\raggedright\arraybackslash}p{\dimexpr 0.23\textwidth-1.0\tabcolsep\relax}>{\raggedright\arraybackslash}p{\dimexpr 0.77\textwidth-1.0\tabcolsep\relax}@{}}
\toprule

\textbf{Tool; registry class} & \textbf{Measured result (two fixtures per listed scenario)} \\

\midrule

git\_diff\_staged; R0 & Both reads return the expected staged change and preserve all measured state. \\
git\_commit; R1 & Ordinary commit + soft reset restores measured state in 2/2, including differing staged and working versions. For pending merges, the same undo loses MERGE\_HEAD, MERGE\_MSG and MERGE\_MODE in 2/2. After a contributor's later commit, HEAD\textasciitilde{}1 removes that later commit from the tip and leaves the agent commit in 2/2; the contributor's file remains on disk. A tip guard defers undo and preserves the later commit in 2/2. \\
git\_reset; R3 & Re-adding a fully staged file restores the staged version in 2/2. For partial staging, re-adding selects the working version rather than the former staged version in 2/2. Working files survive. \\
read\_query; R0 & Both reads return the expected value and preserve logical schema and rows. \\
create\_table; R2 & Ordinary create/drop restores logical state in 2/2. AUTOINCREMENT leaves an empty sqlite\_sequence table in 2/2. With a later inserted row, naive drop erases that row in 2/2 despite matching the original state; a guard defers both drops and preserves the row. \\
write\_query; R4 & Both updates commit before return; a later ROLLBACK has no active transaction. With captured prior values, a conditional reverse update restores the target in 2/2 isolated fixtures and 2/2 fixtures with later edits to another row. If the same row changes later, both reverse updates match zero rows and preserve the later value. \\

\bottomrule
\end{tabular}
\end{minipage}\par\vspace{11pt}
\par\noindent\begin{minipage}{\textwidth}
\small\setlength{\tabcolsep}{4pt}\renewcommand{\arraystretch}{1.22}
\captionof{table}[]{Common tab-close recovery conditions. Each cell represents two draft variants observed twice. No browser had a mismatch between its primary and repeat outcomes. "Empty" describes these recovery routes, not proof of permanent forensic loss.}\label{tab:main7}
\begin{tabular}{@{}>{\raggedright\arraybackslash}p{\dimexpr 0.25\textwidth-1.6\tabcolsep\relax}>{\raggedright\arraybackslash}p{\dimexpr 0.27\textwidth-1.6\tabcolsep\relax}>{\raggedright\arraybackslash}p{\dimexpr 0.16\textwidth-1.6\tabcolsep\relax}>{\raggedright\arraybackslash}p{\dimexpr 0.16\textwidth-1.6\tabcolsep\relax}>{\raggedright\arraybackslash}p{\dimexpr 0.16\textwidth-1.6\tabcolsep\relax}@{}}
\toprule

\textbf{Recovery route} & \textbf{Editor / storage} & \textbf{Chromium} & \textbf{Firefox} & \textbf{Safari} \\

\midrule

Fresh URL & Unsaved textarea & Empty & Empty & Empty \\
Fresh URL & Session storage & Empty & Empty & Empty \\
Fresh URL & Local autosave & Exact & Exact & Exact \\
Fresh URL & Earlier checkpoint & Earlier & Earlier & Earlier \\
Fresh URL & Unsaved rich editor & Empty & Empty & Empty \\
Native tab restore & Unsaved textarea & Exact & Exact & Empty \\
Native tab restore & Session storage & Exact & Exact & Empty \\
Native tab restore & Local autosave & Exact & Exact & Exact \\
Native tab restore & Earlier checkpoint & Exact & Exact & Earlier \\
Native tab restore & Unsaved rich editor & Empty & Empty & Empty \\

\bottomrule
\end{tabular}
\end{minipage}\par\vspace{11pt}
\par\noindent\begin{minipage}{\textwidth}
\small\setlength{\tabcolsep}{4pt}\renewcommand{\arraystretch}{1.22}
\captionof{table}[]{Additional Chromium and Firefox conditions; two variants per cell, repeated. Safari restart and isolated-profile recovery are unmeasured.}\label{tab:main8}
\begin{tabular}{@{}>{\raggedright\arraybackslash}p{\dimexpr 0.3\textwidth-1.6666666666666667\tabcolsep\relax}>{\raggedright\arraybackslash}p{\dimexpr 0.14\textwidth-1.6666666666666667\tabcolsep\relax}>{\raggedright\arraybackslash}p{\dimexpr 0.14\textwidth-1.6666666666666667\tabcolsep\relax}>{\raggedright\arraybackslash}p{\dimexpr 0.14\textwidth-1.6666666666666667\tabcolsep\relax}>{\raggedright\arraybackslash}p{\dimexpr 0.14\textwidth-1.6666666666666667\tabcolsep\relax}>{\raggedright\arraybackslash}p{\dimexpr 0.14\textwidth-1.6666666666666667\tabcolsep\relax}@{}}
\toprule

\textbf{Chromium and Firefox route} & \textbf{No save} & \textbf{Session} & \textbf{Local save} & \textbf{Checkpoint} & \textbf{Rich editor} \\

\midrule

Leave-open control & Exact & Exact & Exact & Exact & Exact \\
Restart, same profile & Empty & Empty & Exact & Earlier & Not tested \\
Restart, isolated profile & Empty & Empty & Empty & Empty & Not tested \\

\bottomrule
\end{tabular}
\end{minipage}\par\vspace{11pt}
\par\noindent\begin{minipage}{\textwidth}
\small\setlength{\tabcolsep}{4pt}\renewcommand{\arraystretch}{1.22}
\captionof{table}[]{Evidence paths relative to the companion archive's data/ directory. Within classes-study, unqualified result filenames in this table are under results/.}\label{tab:main9}
\begin{tabular}{@{}>{\raggedright\arraybackslash}p{\dimexpr 0.27\textwidth-1.0\tabcolsep\relax}>{\raggedright\arraybackslash}p{\dimexpr 0.73\textwidth-1.0\tabcolsep\relax}@{}}
\toprule

\textbf{Evidence directory} & \textbf{Principal records} \\

\midrule

classes-study & results/hint-comparison.csv; replay-summary.csv; replay-by-trace.csv; rule-contracts.csv; private-file-effects.csv; private-db-effects.csv; reviewers/shaad-final-review.csv \\
recovery-study & measurements.csv; run.json; PROTOCOL.txt; run.py \\
git-sqlite-study & measurements.csv; states.json; verify.py; PROTOCOL.md; run.py \\
browser-study & measurements.csv; replication.csv; rich-text.csv; rich-text-repeat.csv; fixture HTML and reproduction instructions \\
cross-browser-study & firefox.csv; firefox-repeat.csv; safari.csv; safari-repeat.csv; safari-ui-raw.json; comparison.csv; verify.py \\

\bottomrule
\end{tabular}
\end{minipage}\par\vspace{11pt}
\clearpage

\subsection{Assisted-review cross-tabulation}

\par\noindent\begin{minipage}{\textwidth}
\small\setlength{\tabcolsep}{4pt}\renewcommand{\arraystretch}{1.22}
\captionof{table}[]{One completed assisted review, all 150 selected items. Rows are draft registry labels; columns are approved ratings. The diagonal contains 75 of the 148 rated items. This is not independent agreement.}\label{tab:review}
\begin{tabular}{@{}>{\raggedright\arraybackslash}p{\dimexpr 0.16\textwidth-1.75\tabcolsep\relax}>{\raggedright\arraybackslash}p{\dimexpr 0.12\textwidth-1.75\tabcolsep\relax}>{\raggedright\arraybackslash}p{\dimexpr 0.12\textwidth-1.75\tabcolsep\relax}>{\raggedright\arraybackslash}p{\dimexpr 0.12\textwidth-1.75\tabcolsep\relax}>{\raggedright\arraybackslash}p{\dimexpr 0.12\textwidth-1.75\tabcolsep\relax}>{\raggedright\arraybackslash}p{\dimexpr 0.12\textwidth-1.75\tabcolsep\relax}>{\raggedright\arraybackslash}p{\dimexpr 0.12\textwidth-1.75\tabcolsep\relax}>{\raggedright\arraybackslash}p{\dimexpr 0.12\textwidth-1.75\tabcolsep\relax}@{}}
\toprule

\textbf{Registry} & \textbf{R0} & \textbf{R1} & \textbf{R2} & \textbf{R3} & \textbf{R4} & \textbf{Unrated} & \textbf{Total} \\

\midrule

R0 & 28 & 0 & 3 & 0 & 3 & 0 & 34 \\
R1 & 0 & 1 & 15 & 10 & 5 & 0 & 31 \\
R2 & 0 & 0 & 5 & 4 & 7 & 2 & 18 \\
R3 & 0 & 0 & 2 & 11 & 20 & 0 & 33 \\
R4 & 1 & 1 & 1 & 1 & 30 & 0 & 34 \\

\bottomrule
\end{tabular}
\end{minipage}\par\vspace{11pt}

\subsection{Replay counts by domain and condition}

\par\noindent\begin{minipage}{\textwidth}
\small\setlength{\tabcolsep}{4pt}\renewcommand{\arraystretch}{1.22}
\captionof{table}[]{Projected approval points from the implemented session guard and read/write proxy. Fewer/equal/more compare approval counts within the same trace. Each recorded run is counted once in this table; repeated task trials remain correlated. Source: \path{classes-study/results/paired-by-domain.csv}.}\label{tab:domain}
\begin{tabular}{@{}>{\raggedright\arraybackslash}p{\dimexpr 0.14\textwidth-1.7777777777777777\tabcolsep\relax}>{\raggedright\arraybackslash}p{\dimexpr 0.13\textwidth-1.7777777777777777\tabcolsep\relax}>{\raggedright\arraybackslash}p{\dimexpr 0.09\textwidth-1.7777777777777777\tabcolsep\relax}>{\raggedright\arraybackslash}p{\dimexpr 0.105\textwidth-1.7777777777777777\tabcolsep\relax}>{\raggedright\arraybackslash}p{\dimexpr 0.105\textwidth-1.7777777777777777\tabcolsep\relax}>{\raggedright\arraybackslash}p{\dimexpr 0.105\textwidth-1.7777777777777777\tabcolsep\relax}>{\raggedright\arraybackslash}p{\dimexpr 0.105\textwidth-1.7777777777777777\tabcolsep\relax}>{\raggedright\arraybackslash}p{\dimexpr 0.11\textwidth-1.7777777777777777\tabcolsep\relax}>{\raggedright\arraybackslash}p{\dimexpr 0.11\textwidth-1.7777777777777777\tabcolsep\relax}@{}}
\toprule

\textbf{Domain} & \textbf{Condition} & \textbf{Runs} & \textbf{Calls} & \textbf{Guard} & \textbf{Write} & \textbf{Fewer} & \textbf{Equal} & \textbf{More} \\

\midrule

banking & benign & 16 & 27 & 8 & 8 & 0 & 16 & 0 \\
banking & injected & 144 & 271 & 76 & 76 & 0 & 144 & 0 \\
slack & benign & 21 & 116 & 47 & 34 & 0 & 12 & 9 \\
slack & injected & 105 & 693 & 304 & 227 & 0 & 57 & 48 \\
travel & benign & 20 & 127 & 6 & 6 & 0 & 20 & 0 \\
travel & injected & 140 & 880 & 42 & 42 & 0 & 140 & 0 \\
workspace & benign & 40 & 101 & 24 & 28 & 4 & 36 & 0 \\
workspace & injected & 560 & 1471 & 343 & 399 & 56 & 504 & 0 \\
airline & benign & 200 & 1678 & 332 & 332 & 0 & 200 & 0 \\
retail & benign & 456 & 3591 & 776 & 776 & 0 & 456 & 0 \\
telecom & benign & 456 & 3121 & 711 & 711 & 0 & 456 & 0 \\

\bottomrule
\end{tabular}
\end{minipage}\par\vspace{11pt}

\clearpage

\subsection{Attributable positive runs by domain}

\par\noindent\begin{minipage}{\textwidth}
\small\setlength{\tabcolsep}{4pt}\renewcommand{\arraystretch}{1.22}
\captionof{table}[]{Attribution and policy coverage for attack-positive recorded runs. Zero attributed cases is absence of attributable evidence, not proof of successful prevention. Source: \path{classes-study/results/replay-summary.csv}.}\label{tab:attack-domain}
\begin{tabular}{@{}>{\raggedright\arraybackslash}p{\dimexpr 0.16\textwidth-1.7142857142857142\tabcolsep\relax}>{\raggedright\arraybackslash}p{\dimexpr 0.14\textwidth-1.7142857142857142\tabcolsep\relax}>{\raggedright\arraybackslash}p{\dimexpr 0.14\textwidth-1.7142857142857142\tabcolsep\relax}>{\raggedright\arraybackslash}p{\dimexpr 0.14\textwidth-1.7142857142857142\tabcolsep\relax}>{\raggedright\arraybackslash}p{\dimexpr 0.14\textwidth-1.7142857142857142\tabcolsep\relax}>{\raggedright\arraybackslash}p{\dimexpr 0.14\textwidth-1.7142857142857142\tabcolsep\relax}>{\raggedright\arraybackslash}p{\dimexpr 0.14\textwidth-1.7142857142857142\tabcolsep\relax}@{}}
\toprule

\textbf{Domain} & \textbf{Injected runs} & \textbf{Positive goals} & \textbf{Attributed runs} & \textbf{Guard gated} & \textbf{Write gated} & \textbf{Severity gated} \\

\midrule

banking & 144 & 6 & 6 & 6 & 6 & 6 \\
slack & 105 & 25 & 14 & 14 & 3 & 0 \\
travel & 140 & 1 & 0 & 0 & 0 & 0 \\
workspace & 560 & 15 & 9 & 9 & 9 & 3 \\

\bottomrule
\end{tabular}
\end{minipage}\par\vspace{11pt}

\subsection{Git and SQLite scenario-level results}

\par\noindent\begin{minipage}{\textwidth}
\small\setlength{\tabcolsep}{4pt}\renewcommand{\arraystretch}{1.22}
\captionof{table}[]{Six tools and 16 scenarios, with ASCII and Unicode fixtures per scenario (32 primary observations). Exact equality uses the state boundary in Section 6.6. Matching initial state can destroy later work, as shown by the later-insert drop. N/A means there was no later writer. Source: \path{git-sqlite-study/measurements.csv}.}\label{tab:git-scenarios}
\begin{tabular}{@{}>{\raggedright\arraybackslash}p{\dimexpr 0.18\textwidth-1.6\tabcolsep\relax}>{\raggedright\arraybackslash}p{\dimexpr 0.23\textwidth-1.6\tabcolsep\relax}>{\raggedright\arraybackslash}p{\dimexpr 0.24\textwidth-1.6\tabcolsep\relax}>{\raggedright\arraybackslash}p{\dimexpr 0.16\textwidth-1.6\tabcolsep\relax}>{\raggedright\arraybackslash}p{\dimexpr 0.19\textwidth-1.6\tabcolsep\relax}@{}}
\toprule

\textbf{Tool} & \textbf{Scenario} & \textbf{Recovery outcome} & \textbf{Initial state matched} & \textbf{Later work preserved} \\

\midrule

\nolinkurl{git_diff_staged} & read & read preserved state & 2/2 & N/A \\
\nolinkurl{git_commit} & ordinary commit & soft reset & 2/2 & N/A \\
\nolinkurl{git_commit} & merge commit & soft reset & 0/2 & N/A \\
\nolinkurl{git_commit} & later commit reset & soft reset & 0/2 & 0/2 \\
\nolinkurl{git_commit} & later commit guard & deferred & 0/2 & 2/2 \\
\nolinkurl{git_reset} & fully staged & restaged original & 2/2 & N/A \\
\nolinkurl{git_reset} & partially staged & staged version differs & 0/2 & N/A \\
\nolinkurl{read_query} & read & read preserved state & 2/2 & N/A \\
\nolinkurl{create_table} & plain table & dropped table & 2/2 & N/A \\
\nolinkurl{create_table} & autoincrement table & dropped table & 0/2 & N/A \\
\nolinkurl{create_table} & later insert drop & dropped table & 2/2 & 0/2 \\
\nolinkurl{create_table} & later insert guard & deferred & 0/2 & 2/2 \\
\nolinkurl{write_query} & rollback after return & no open transaction & 0/2 & N/A \\
\nolinkurl{write_query} & captured alone & restored target & 2/2 & N/A \\
\nolinkurl{write_query} & later other row & restored target & 0/2 & 2/2 \\
\nolinkurl{write_query} & later same row & deferred & 0/2 & 2/2 \\

\bottomrule
\end{tabular}
\end{minipage}\par\vspace{11pt}

\subsection{Measurement-file inventory}

Table~\ref{tab:inventory} lists the primary and repeated measurements and the main derived summaries, with counts read from the packaged files. CSV counts exclude the header. The replay-by-trace table repeats each of 2,158 traces across seven policies; the 15,106 rows are not 15,106 independent runs. The Git/SQLite primary and repetition measurements and complete fixture states are included. The repeated CSV and state JSON are byte-identical to the primary files. Browser repetitions are separately packaged. Full public benchmark traces and upstream source checkouts are retrieved using the supplied scripts and are not embedded here.

\par\noindent\begin{minipage}{\textwidth}
\small\setlength{\tabcolsep}{4pt}\renewcommand{\arraystretch}{1.22}
\captionof{table}[]{Inventory of core observation and summary files. Hashes are provided in evidence-inventory.json and the archive-wide SHA256SUMS.json. Blank independent-review templates are not counted as completed reviews.}\label{tab:inventory}
\begin{tabular}{@{}>{\raggedright\arraybackslash}p{\dimexpr 0.54\textwidth-1.3333333333333333\tabcolsep\relax}>{\raggedright\arraybackslash}p{\dimexpr 0.1\textwidth-1.3333333333333333\tabcolsep\relax}>{\raggedright\arraybackslash}p{\dimexpr 0.36\textwidth-1.3333333333333333\tabcolsep\relax}@{}}
\toprule

\textbf{File relative to data/} & \textbf{Records} & \textbf{Role} \\

\midrule

\nolinkurl{classes-study/reviewers/shaad-final-review.csv} & 150 & Assisted decisions; two unrated \\
\nolinkurl{classes-study/inputs/replay.jsonl} & 2158 & Recorded-run metadata \\
\nolinkurl{classes-study/results/hint-comparison.csv} & 536 & Tool-level hint comparison \\
\nolinkurl{classes-study/results/hint-confusion.csv} & 5 & Derived class cross-tabulation \\
\nolinkurl{classes-study/results/approval-by-server.csv} & 20 & Derived approval comparisons \\
\nolinkurl{classes-study/results/replay-tool-mappings.csv} & 106 & Analyst tool adapters \\
\nolinkurl{classes-study/results/replay-summary.csv} & 91 & Derived policy/domain summaries \\
\nolinkurl{classes-study/results/replay-by-trace.csv} & 15106 & Trace-policy projections \\
\nolinkurl{classes-study/results/paired-by-domain.csv} & 11 & Derived paired comparisons \\
\nolinkurl{classes-study/results/goal-positive-runs.csv} & 47 & Positive-label attribution records \\
\nolinkurl{classes-study/results/replay-private-capture.csv} & 59 & Capture eligibility and recovery \\
\nolinkurl{classes-study/results/rule-contracts.csv} & 738 & Rule-case checks \\
\nolinkurl{classes-study/results/private-file-effects.csv} & 6 & Initial file effect probes \\
\nolinkurl{classes-study/results/private-db-effects.csv} & 3 & Initial PostgreSQL probes \\
\nolinkurl{recovery-study/measurements.csv} & 48 & Later-writer policy fixtures \\
\nolinkurl{git-sqlite-study/measurements.csv} & 32 & Primary MCP observations \\
\nolinkurl{git-sqlite-study/states.json} & 32 & Full measured fixture states \\
\nolinkurl{git-sqlite-study/replication/measurements.csv} & 32 & Repeated MCP observations \\
\nolinkurl{git-sqlite-study/replication/states.json} & 32 & Repeated full fixture states \\
\nolinkurl{browser-study/measurements.csv} & 40 & Primary textarea observations \\
\nolinkurl{browser-study/replication.csv} & 40 & Repeated textarea observations \\
\nolinkurl{browser-study/rich-text.csv} & 6 & Primary rich-editor observations \\
\nolinkurl{browser-study/rich-text-repeat.csv} & 6 & Repeated rich-editor observations \\
\nolinkurl{cross-browser-study/firefox.csv} & 46 & Primary Firefox observations \\
\nolinkurl{cross-browser-study/firefox-repeat.csv} & 46 & Repeated Firefox observations \\
\nolinkurl{cross-browser-study/safari.csv} & 20 & Primary Safari observations \\
\nolinkurl{cross-browser-study/safari-repeat.csv} & 20 & Repeated Safari observations \\
\nolinkurl{cross-browser-study/safari-ui-raw.json} & 40 & Raw Safari cases; two page observations each \\
\nolinkurl{cross-browser-study/comparison.csv} & 10 & Derived common-route comparison \\

\bottomrule
\end{tabular}
\end{minipage}\par\vspace{11pt}

\clearpage

\subsection{Reproduction, provenance and availability}

The evidence files accompany this manuscript as a local research bundle; this draft does not assert that they have already been deposited in a public repository or assigned a DOI. Before public submission, the authors should select a stable artifact location and replace this availability statement with its persistent identifier. Existing public source-code and benchmark citations remain in the references.

For the core study, \path{classes-study/REPRODUCE.txt} documents source retrieval, classifier execution, annotation extraction, replay and effect checks. The later-writer file study includes \path{recovery-study/PROTOCOL.txt}, \path{run.py} and \path{run.json}; the latter records the measurement and protocol hashes. Git/SQLite and browser studies include their own protocols, verification scripts and reproduction instructions. Safari's actual native-UI procedure and its deviations from the initial driver plan are documented in \path{cross-browser-study/SAFARI-PROTOCOL.md} and \path{RESULTS.md}. Unique Safari URLs and keys were used in the same profile and loopback origin, rather than independent origins.

The data checks can be run from the extracted archive root as follows. These commands validate saved observations; they do not execute fresh browser experiments. The cross-browser verifier regenerates the same derived Safari CSVs and common comparison from recorded observations.

\begin{quote}\small\ttfamily
python data/paper/verify\_metrics.py\\
python data/git-sqlite-study/verify.py data/git-sqlite-study\\
python data/browser-study/verify.py\\
python data/cross-browser-study/verify.py
\end{quote}

The LaTeX manuscript uses standard packages and six original vector PDF figures in \path{figures/}. Figure-generation code and source-data hashes are included. Compile \path{main.tex} with Tectonic, or with a standard LaTeX installation using two pdfLaTeX passes to resolve citations and cross-references. The source archive includes the compiled PDF, build instructions, provenance and an integrity manifest. All numerical figure inputs are also retained as tables and machine-readable data. The overview and recovery schematics illustrate declared mechanisms and measured fixture sequences; they are not additional experiments.

\end{document}